# Combining Linguistic and Spatial Information for Document Analysis

Marco Aiello[1,2], Christof Monz[1] and Leon Todoran[2]
[1]Institute for Logic, Language and Computation (ILLC)
[2]Intelligent Sensory and Information Systems (ISIS)
University of Amsterdam, The Netherlands
E-mail: {aiellom,christof,todoran}@science.uva.nl

**Abstract**

We present a framework to analyze color documents of complex layout. In addition, no assumption is made on the layout. Our framework combines in a content-driven bottom-up approach two different sources of information: textual and spatial. To analyze the text, shallow natural language processing tools, such as taggers and partial parsers, are used. To infer relations of the logical layout we resort to a qualitative spatial calculus closely related to Allen's calculus. We evaluate the system against documents from a color journal and present the results of extracting the reading order from the journal's pages. In this case, our analysis is successful as it extracts the intended reading order from the document.

**Introduction**

The idea behind the first attempts to create digital libraries from paper documents, was to archive directly the scanned documents. Besides the problem of huge memory requirements, the process of retrieval was difficult and unsatisfactory, especially because the indexing method was based on the file name and nothing else. The next step was to manually annotate document images, and use the annotation to implement the retrieval. But manual annotation requires a lot of human resources and is feasible only for small-sized databases. Nowadays document analysis tries to automatically extract information from document images.

Document analysis can also be viewed as the reverse process of document authoring. Therefore, the result of document analysis is expected to be the same document representation as the one used in document creation. There are various formats used for document representation, and standards like ODA and SGML try to unify document representations. Mainly two distinct structures used in document representation can be distinguished:

- the *layout structure,* which precisely describes how the document looks like, and

- the *logical structure,* which gives an interpretation or understanding of the document.

The layout structure of a document is a set of document objects (as characters, words, text lines, text boxes, pictures) with their features (like position, bounding box, color, font information) and relations between the objects (e.g. adjacency, overlapping, horizontal-sequence, part-of). The set of document objects and the set of relations have redundant items. The choice of document objects and relations to be used depends on the document analysis application and document class considered. For instance Tsujimoto and Asada (1992) use a description at the block level, and a tree structure representing the vertical/horizontal arrangement of blocks in page. Worring and Todoran (1999) use a directed graph structure to represent the overlapping relations between blocks in color documents.

The analysis of the logical structure interprets the document objects as found in the document's layout, and establishes logical relations among them. For example, some textual document objects will be labeled as title, section, page number, etc. Pictures can be classified as drawings, images, charts, etc. The most common relation considered among logical document objects is the reading order; but some other relations can also be employed. For example the document hierarchy, or specific relations between document parts are used in technical drawing analysis.

Several methods for layout detection can be found in the literature. *Bottom-up* approaches start with pixel information and sequentially group the detected entities up to the document objects. *Top-down* approaches consider a document model as the pattern to reveal the layout structure of document instances. Almost all published methods are for binary or gray level images, and just few attempts have been made to process color document. One can say that the problem of layout detection is solved for black-and-white documents, having a regular layout. The challenge remains for complex (color) documents. Presently, the most difficult task in document analysis is to extract the logical structure given the layout. Current approaches in literature use a priori information about the document's class. There are efficient methods working on specific classes of documents, but they cannot be applied on generic documents. Tsujimoto and Asada (1992) uses a tree representation for layout of multi-column black-and-white documents. A conversion from layout structure to logical structure is made in two steps. First a set of 4 transformation rules (based on geometrical principles) is used to convert the layout tree into a logical tree which represents the reading order. Then another set of rules is used to classify the tree nodes as instances of title, abstract, caption, subtitle, paragraph, header, footer and page number.

Ishitani (1999) uses a similar method exploiting geometrical properties of a given class of documents. The transformation rules encode a priori information about the document's class. Rather than having two disconnected steps: layout and logical structure detection, Ishitani exchanges information between layout and logical analysis which are applied iteratively. The results are improved both in layout and logical detection.

In this paper, we discuss the detection of logical structure given the layout structure by combining different kinds of knowledge extracted from the document: spatial, lexico-graphic and grammatical. We test our framework by trying to establish the reading order. In contrast with previous methods, we take full advantage of the text present in the document as we use it by a Natural Language Processing (NLP) component to reconstruct the logical structure of the document.

**The Architecture**

The analysis of a color document image begins by the extraction of the layout and the identification of the document objects. This is achieved by an existing tool (Worring & Todoran, 1999). The layout information is then passed on to the spatial inference component, which identifies the complex spatial relations among the document objects and gives a number of possible readings of the document. A module of the spatial inference component stores the logical rules that are used in the analysis of the document at hand. These rules can be either very general or document dependent. Finally, the possible document readings with the text present in the textual document objects are passed to the NLP component which attempts to further disambiguate the possible readings. The output of the system is a set of possible readings of the document. We call the analysis *fully successful* if the set is composed by one single element which coincides with the order intended by the document author(s).

**Exploiting automatic spatial inference**

Once the document objects and their spatial arrangement have been identified, the next step is to search for the logical structure of the document. In other words, how did the author(s) of the document arrange the items to convey information to the reader? Some techniques are quite general for documents (in the occidental culture, it is common knowledge that the reading order is from left to right and from top to bottom within a textual unit), other are more specific to a document category or even to specific document instances (e.g., some journals impose the picture to be only at the beginning of a page). In this section, we describe how one can exploit spatial inference to automatically process the available layout data, in the attempt to reconstruct the logical structure of the document. For our purposes, it is convenient and sufficient to consider the document objects as rectangles, i.e. to consider the bounding boxes of the objects.

The formal tool we are using to perform spatial reasoning—in a qualitative manner—is the rectangle model as defined in (Balbiani, Condotta & Fariñas del Cerro, 1998) . More precisely, a rectangle model is a triple $\langle R, m_x, m_y \rangle$, where $R$ is a set of rectangles and $m_x$ and $m_y$ are two sets of basic interval relations holding on the axis among the rectangles $R$. Each set $m_x, m_y$ consists of 13 exhaustive and pairwise disjoint relations of the the form: `precedes, meets, overlaps, starts, during, finishes, equals` (and their inverses). Such relations were originally defined by Allen (1983) (and independently by van Benthem (1983)) to devise a complete calculus for time intervals. Since in the rectangle model the relation that holds between any two given rectangles is the product of a relation on the *x*-axis and one on the *y*-axis, there are exactly $13^2$ rectangle relations. These relations form a complete calculus on the rectangles for which compositional tables are available (cf. Allen, 1983). Moreover, given any two rectangles, there is always one and only one of the $13^2$ relations that holds among them.

In the specific context of document analysis, the layout detection is used to make an instance of a rectangle model, encoding the document objects as rectangles in $R$ and computing the basic relations $m_x, m_y$ among them. At the same time, the logical properties of the document are available as complex relations (built from the basic $m_x, m_y$). Now, there are two ways to proceed. One is to consider all relations between document objects which are consistent with the document encoding. This is theoretically a quite complex task to perform, since it is of exponential complexity in the number of document objects. Though in practice, the number of document objects is quite small in most cases and for a great variety of documents. The second and more lightweight way to proceed is to use the rectangle model for disambiguation. Take any kind of result of the analysis of the document, e.g. the reading order. One can match it against the rectangle model to see if it is consistent. To this end, one transforms the reading order into a set of constraints holding among document objects. For example, if A1 should be read before A2, then `precedes_on_x(A1,A2)` or `precedes_on_y(A1,A2)` are added to a set of constraints. Then, one checks whether the constraints are consistent with the rectangle model. This consistency checking is encoded as a constraint satisfaction problem. In (Balbiani, Condotta & Fariñas del Cerro, 1998), it is shown that this problem is decidable and the complexity has a polynomial bound.

**Exploiting Language Technology**

One of the aims of document analysis approaches is to reconstruct the reading order of a document. Astonishingly, in previous attemps this is done without considering textual information at all. Complex layout structures using multiple frames, overlapping frames, and discontinuous texts (e.g., "continuation from page 73"), make it very difficult to construct the reading order just by considering geometric information.

Inevitably, situations arise where several reading orders are possible. For instance, consider the document in Figure 1. As shown in Figure 1 (c), two reading orders are possible: (1; 2; 6; 7) or (1; 6; 2; 7), where 1, 2, 6, and 7 are the text blocks of the document.

For short, end(1) and beg(2) represent the last and the first sentence fragment of block 1 and 2, respectively. To decide whether (1;2) guarantees a correct reading order, we check whether concatenating end(1) and beg(2), i.e., end(1) ∘ beg(2), results in a well-formed sentence, unless end(1) ends with a period. In practice, even for a large collection of documents, feasibility can be accomplished by applying CASS, (cf. Abney, 1990), a robust and fast finite state parser to end(1) ∘ beg(2). Since CASS requires its input to be tagged (i.e., that all words are annotated with their part-of-speech), LTPOS (cf. Mikheev 1997), a part-of-speech tagger is applied to the text blocks first. Table 1 shows an instance of this problem. Here, both, end(1) ∘ beg(2) and end(1) ∘ beg(6), are checked for well-formedness, but end(2) ∘ beg(6) does not lead to a well-formed sentence, which allows us to reject (1; 2; 6; 7) as a possible reading order. On the other hand, end(1) ∘ beg(6), end(6) ∘ beg(2), and end(2) ∘ beg(7) can be parsed as well-formed sentences, and (1;6;2;7) is considered to be a possible reading order. Note, that these decisions can be made without relying on any meta-information on the document layout format. The only assumption we make is that we know which natural language the document is written in.

|   | Begin | End |
|---|---|---|
| 1 |  | The CLASS attribute can hold information that would otherwise be lost when converting a document to HTML, and a style sheet can act |
| 2 | but HTML lacks special elements for mathematics. | Instead, a document can claim to be well-formed by following some simple syntactical rules. |
| 6 | on the value of the CLASS attribute (see Figure 2). | For example mathematicians may want to encode formulae inside HTML documents, |
| 7 | The XML specification became a W3C Recommendation in February 1998, and its first uses have appeared [1]. |  |

Table 1: Sentence fragments beginning or ending a paragraph.

In general, when considering possible concatenations of the form end($m$) ∘ beg($n$), three situations can be distinguished:

1. $m$ ends with a hyphenated word (e.g., 'prod-')
2. $m$ ends with a word
3. $m$ ends with some kind of punctuation (e.g., ',', ';', '.', '!', etc.)

To disambiguate the reading order in these cases, three different NLP tools are applied. If $m$ ends with a hyphenated word $w$, we simply concatenate $w$ with the first word of $n$ and check by means of a lexical look-up whether the the resulting word is well-formed, i.e. in the lexicon. For the lexicon the look-up, we use the lexicon generated from the Brown and Wall-Street-Journal corpora.

To distinguish the second case from the third, it is necessary to recognize the difference between punctuation that is part of a word, as in abbreviations (for instance, 'approx.' or 'e.g.') or acronyms (for instance 'U.S.') and punctuation that marks a sentence boundary. This is accomplished by using a tokenizer (see Schmid 1994) which checks words against a list of abbreviations and acronyms and applies several heuristics to recognize sentence boundaries. If the last word of $m$ does not mark a sentence boundary, end($m$) ∘ beg($n$) is parsed, as described above.

If *m* ends with a punctuation mark, it is often not easy to decide by using shallow NLP tools whether block *m* can immediately proceed block *n* to constitute a possible reading order. Obviously, *m* immediately preceding *n* can be excluded if the first word of *n* starts with a lower-case letter. If on the other hand it starts with an upper-case letter, it is impossible to decide on purely syntactic grounds whether *m* followed by *n* constitutes a possible reading order. It would be necessary to give a deeper semantic analysis to check whether these sentences are semantically related to each other. Tackling this task in a robust, efficient and general way is far beyond the current state of the art in language technology.

**Experimental Results**

We have applied our method to images from the http://www.acm.org/cacmCommunications of the Association for Computing Machinery (CACM) and to the http://www.mediateam.oulu.fi/MTDB/index.htmlMedia Team Document Database II (MTDB). The CACM is a color journal with text in English and with a quite elaborate layout. We have concentrated on the extraction of the reading order. To this end, in the spatial inference component (implemented in http://www-icparc.doc.ic.ac.uk/eclipseEclipse , a prolog based constraint satisfaction problem solver) we have used a simple encoding: "a text block is before another one if it is before on either axis":

```
before_in_reading(B1, B2):-         before_in_reading(B1, B2):-
     precedes_X(B1, B2).                  precedes_Y(B1, B2).

before_in_reading(B1, B2):-         before_in_reading(B1, B2):-
     meets_X(B1, B2).                     meets_Y(B1, B2).

before_in_reading(B1, B2):-         before_in_reading(B1, B2):-
     overlaps_X(B1, B2).                  overlaps_Y(B1, B2).
```

As simple as it may seem, this encoding has proved to be quite powerful to extract all geometrically admissible reading orders from the documents. Consider the following example: a page from the CACM (as depicted in Figure 1).

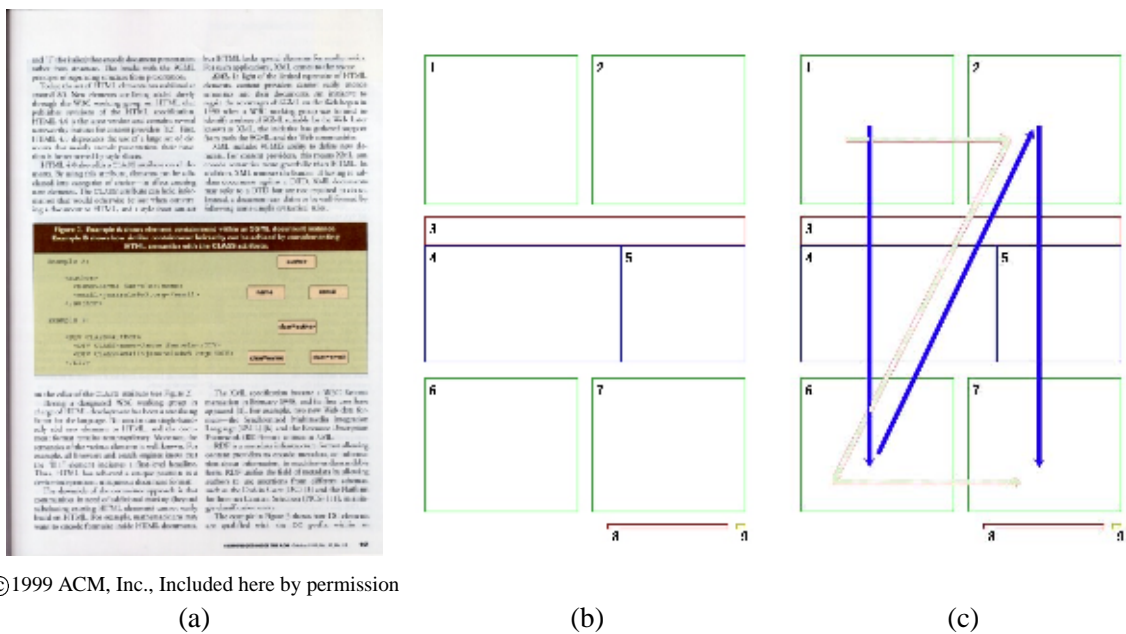

©1999 ACM, Inc., Included here by permission
        (a)                                 (b)                              (c)

Figure 1: The analysis of a CACM page. From left to right in the figure we have (a) the scanned image, (b) the document's layout, and (c) the results of the analysis.

By means of the layout detection component, we are able to extract the geometrical structure of the document (central image) together with all the layout properties of the blocks (type of block, color of background, coordinates, color, size and font type of the possibly present text, etc.).

```
[1, 1, [13, 23, 93, 101],   TimesNewRoman , 11, 0, 16777215]
[2, 1, [100, 23, 180, 101], TimesNewRoman , 11, 0, 16777215]
[3, 2, [13, 107, 180, 122], ArialBold , 11, 0, 16777215]
[4, 8, [13, 122, 115, 183], Courier , 11, 16777215, 0]
[5, 9, [115, 122, 180, 183], None , 11, 0, 16777215]
[6, 1, [13, 191, 93, 261],  TimesNewRoman , 11, 0, 16777215]
[7, 1, [100, 191, 180, 261], TimesNewRoman , 11, 0, 16777215]
[8, 2, [108, 267, 171, 270], ArialBold , 7, 0, 16777215]
[9, 13, [175, 267, 180, 270], ArialBold , 12, 0, 16777215]
```

This information is passed on to the spatial inference component, which calculates all admissible relations holding among the blocks (note that if one relation is not present it means that its inverse holds!).

```
[1, 2], [1, 6], [1, 7], [2, 6], [2, 7], [6, 2], [6, 7]
```

¿From this analysis it is possible to decide which are all the possible reading orders, in our example:

```
[1, 6, 2, 7]        [1, 2, 6, 7]
```

In Figure 1 the first is represented by black arrows and the second by gray arrows (blue and gray respectively, if you have access to the color version of this article). Finally, the possible orders, together with the textual contents of the blocks are passed to the natural language component, which disambiguates further, yielding as unique reading order

```
[1, 6, 2, 7]
```

Let us now consider a more elaborate example: a couple of pages from the CACM (see Figure 2).

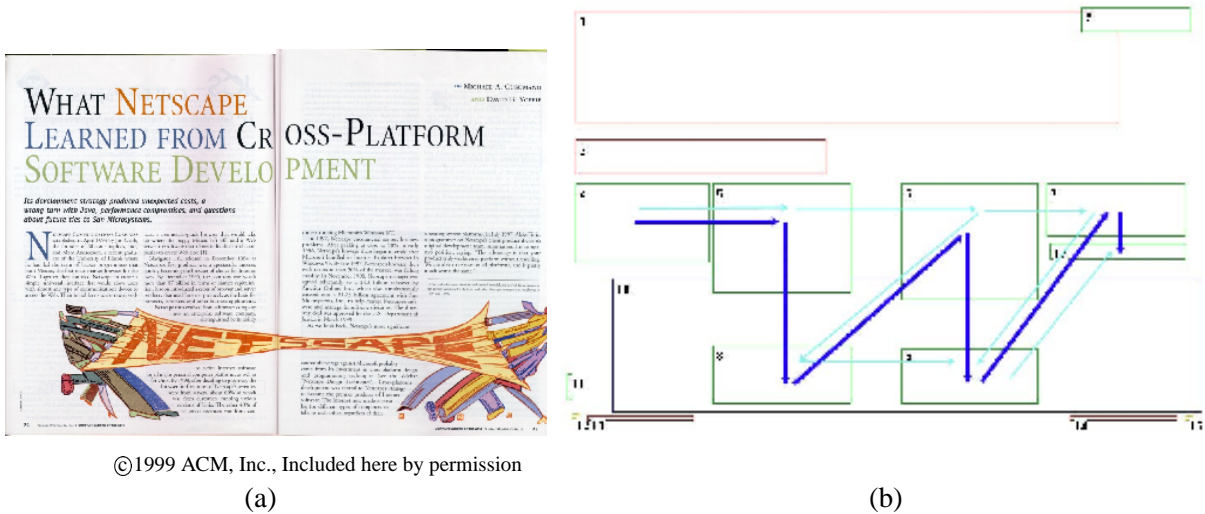

©1999 ACM, Inc., Included here by permission
(a) (b)

Figure 2: The analysis of two CACM pages. From left to right in the figure we have (a) the scanned image, and (c) the results of the analysis.

The analysis proceeds as for the case depicted in Figure 1. The layout is extracted from the two pages, and the following document objects are identified:

```
[1, 2, [20, 25, 345, 93],    TimesNewRoman , 24, 0, 16777215]
[3, 2, [20, 102, 170, 123],  ArialBold     , 11, 0, 16777215]
[4, 1, [20, 128, 100, 176],  TimesNewRoman , 11, 16777215, 0]
[5, 1, [102, 128, 185, 194], TimesNewRoman , 11, 0, 16777215]
[6, 1, [215, 128, 297, 198], TimesNewRoman , 11, 0, 16777215]
[7, 1, [302, 128, 385, 160], TimesNewRoman , 11, 0, 16777215]
[8, 1, [102, 225, 185, 260], TimesNewRoman , 7, 0, 16777215]
[9, 1, [215, 227, 297, 260], TimesNewRoman , 12, 0, 16777215]
[10, 3, [42, 185, 395, 265], None          , 12, 0, 16777215]
[12, 13, [18, 266, 22, 269], TimesNewRoman , 12, 0, 16777215]
[13, 2, [27, 267, 91, 270],  TimesNewRoman , 12, 0, 16777215]
[14, 2, [316, 267, 380, 270], TimesNewRoman , 12, 0, 16777215]
[15, 13, [385, 267, 390, 270], TimesNewRoman , 12, 0, 16777215]
[16, 5, [304, 162, 385, 162], TimesNewRoman , 12, 0, 16777215]
[17, 1, [304, 164, 385, 174], TimesNewRoman , 12, 0, 16777215]
```

The spatially admissible relations among the text blocks are (again, note that if a relation is not present its inverse holds):

```
[4, 5], [4, 6], [4, 7], [4, 8], [4, 9], [4, 17], [5, 6], [5, 7], [5, 8],
[5, 9], [5, 17], [6, 7], [6, 8], [6, 9], [6, 17], [7, 8], [7, 9], [7, 17],
[8, 6], [8, 7], [8, 9], [8, 17], [9, 7], [9, 17], [17, 8], [17, 9]
```

¿From this set of relations we can infer that the possible reading orders are:

```
[4, 5, 8, 6, 9, 7, 17]    [4, 5, 8, 6, 7, 17, 9]    [4, 5, 8, 6, 7, 9, 17]
[4, 5, 6, 8, 9, 7, 17]    [4, 5, 6, 8, 7, 17, 9]    [4, 5, 6, 8, 7, 9, 17]
[4, 5, 6, 7, 17, 8, 9]    [4, 5, 6, 7, 8, 17, 9]    [4, 5, 6, 7, 8, 9, 17]
```

Finally, the language component eliminates the grammatically incorrect and misspelled cases, leaving as unique admissible reading order:

```
[4, 5, 8, 6, 9, 7, 17]
```

We have tested our system on a number of pages from the CACM. In the table below, we display some results from our testing. Each entry in the table has the reference to the CACM issue and page, the number of document objects detected (#Bl), the number of blocks of running text (#Txt_Bl), the number of possible readings of the blocks (#Poss_r), the number of spatially admissible readings (#Spat_admiss_r), i.e. the output of the spatial inference component, the number of final readings of the system (#Final), i.e. the output of the natural language component, and we remark with Correct, if the actual reading order is present in the reading orders given by our system.

| Reference | #Bl | #Txt_Bl | #Poss_r | #Spat_admiss_r | #Final | Correct |
|---|---|---|---|---|---|---|
| CACMv42n10p91 | 9 | 4 | 24 | 1 | 1 | √ |
| CACMv42n11p72 | 17 | 7 | 5040 | 9 | 1 | √ |
| CACMv42n11p97 | 9 | 4 | 24 | 2 | 1 | √ |
| CACMv42n12p20 | 9 | 5 | 120 | 2 | 1 | √ |

Table 2: Results of applying the framework to pages from the CACM.

We have also tested the spatial inference component on the MTDB. We have only used the spatial inference component since the documents come already with the layout information preprocessed and the text is in different natural languages. In this case, the results of our testing have been the following, where Ex_t is the execution time in seconds on a Sun Ultra-5_10 with 64Mb RAM:

| Reference | #Bl | #Txt_Bl | #Poss_r | #Spat_admiss_r | Ex_t | Correct |
|---|---|---|---|---|---|---|
| AR00225_P00531 | 11 | 5 | 120 | 4 | 0.04 | √ |
| AR00233_P00550 | 37 | 12 | $4.8 \cdot 10^8$ | 396 | 3.5 | √ |
| AR00365_P00939 | 9 | 4 | 24 | 1 | 0.01 | √ |
| AR00385_P00991 | 30 | 11 | $3.99 \cdot 10^7$ | 165 | 1.41 | √ |
| average on 206 doc | 10.9709 | 2.82524 | 4.83871 | 1.99515 | 0.0114563 | √ |

Table 3: Results of applying the spatial inference component to the MTDB.

In particular for 137 documents of the set of 206 the reading order is uniquely identified; while 7 give more than 100 spatially admissible readings. A question of effectiveness of the method thus arises. Which documents are effectively and usefully analyzed by the spatial inference component? We measure the utility of applying our method by the following formula:

$$\text{utility} = \frac{\sum(\frac{\#Spat\_admiss}{\#Poss\_r} \cdot \frac{1}{Correct})}{\text{number of documents}}$$

where the sum is intended over all documents analyzed and correct is either 1 or 0 depending on the fact that the intended reading order is in the set of the spatially admissible ones or not, respectively. The smaller the value of the utility measure the more valuable is the application of the system. Values range from 1 to 0, and go to infinity if the correct answer is not present in the set returned by the system. We can now measure the utility of applying our method to the CACM in 0.1434, while for the 206 documents of MTDB it is 0.4123. In the case of big collections of documents with very different layouts, like the MTDB, it is more appropriate to consider an utility measure based on the median value rather than the average. Using $ut_m = \text{median}(\frac{\#Spat\_admiss}{\#Poss\_r} \cdot \frac{1}{Correct})$, we get 0.1667 for the 206 documents of the MTDB.

**Discussion**

We have proposed the use of spatial inference and natural language components to enhance the extraction of logical structure from color documents. We have tested our framework having as goal the reconstruction of the reading order of running English text. Though, the framework is more widely applicable. The geometric rules that drive the spatial inference can be enriched to capture also the spatial behavior of all kinds of document objects, and therefore reconstruct the logical structure of entire documents.

Many of the short-comings of previous approaches, e.g. the fundamental (Tsujimoto & Asada, 1992), are overcome by our framework. Unlike Tsujimoto and Asada, we handle very different types of documents, allowing color to be used, dealing with overlapping and nested document objects, and considering logical structure with independent paths. In other words, we do not commit to tree-like representations neither to express the layout information or the logical information. This because tree-like representations cannot capture many kinds of relevant topological situations without making a choice that is not present *per se* in the document. A tree-like representation forces an order between document objects that the document does not necessarily suggest. Moreover, the properties of our algorithm are known and facts can be proved about its behavior and complexity (by building on the results in Balbiani, Condotta & Fariñas del Cerro, 1998), while in (Tsujimoto & Asada, 1992) nothing can be said about the structure extracting algorithm.

Finally, our approach is highly modular in that the encoding of the logical structure in terms of the rectangle model can be varied quite easily and the whole logical analysis step still goes through. Consider again the CACM documents, if we add three simple document dependent predicates, such as

```
before_in_reading(B1, B2):-
      precedes_Y(B1, B2), before_in_reading_on_X(B1, B2).

before_in_reading(B1, B2):-
      meets_Y(B1, B2), before_in_reading_on_X(B1, B2).

before_in_reading(B1, B2):-
      overlaps_Y(B1, B2), before_in_reading_on_X(B1, B2).
```

to rule out the possibility of reading a text block B1, before a text block B2 which is above and to the right, the whole extraction of the reading order still works. Even more interestingly, the spatial inference component is now able to uniquely spot the right reading order for all CACM pages (which makes the value of the utility measure, defined in the previous section, drop to 0.023).

As we plan to experiment and investigate further some concerns arise from the possibility of exponential blow-ups as the number of document object increases. If this will prove to be the case, we will research the feasibility of applying locality principles in the extraction of the logical structure. This would also give a notion of proximity, which is not expressible within the rectangle model (cf. Singh, Lahoti & Mukerjee, 1999).

**Acknowledgments**

The sample documents used in this article are scanned from the Communications of the ACM, volume 42, number 10, pages 97, 72 and 73. The ACM is acknowledged for kindly giving us permission to reproduce the pages.

Marco Aiello is partially funded by CNR (grant 203.15.10, 11/06/98), Christof Monz was supported by the Physical Sciences Council with financial support from the Netherlands Organization for Scientific Research (NWO), project 612-13-001.